\documentclass[preprint,12pt]{elsarticle}

\usepackage{amssymb}
\usepackage{amsmath}
\usepackage{hyperref}
\usepackage[utf8]{inputenc} 
\usepackage[numbers]{natbib}
\usepackage{xspace}
\usepackage{pgfplots}
\usepackage{multirow}
\usepackage{caption}
\usepackage{subcaption}

\newcommand{\eg}{\emph{e.g.,}\xspace}

\journal{Artificial Intelligence in Medicine}

\begin{document}

\begin{frontmatter}

\title{Exploring the Feasibility of AI-Assisted Spine MRI Protocol Optimization Using DICOM Image Metadata}

\author[inst1]{Alice Vian}
\ead{abvian@inf.ufrgs.br}

\affiliation[inst1]{department={Institute of Informatics},  organization={Universidade Federal do Rio Grande do Sul (UFRGS)},
            addressline={Av. Bento Gonçalves, 9500}, 
            city={Porto Alegre},
            postcode={91501-970}, 
            state={Rio Grande do Sul},
            country={Brazil}
            }

\author[inst2]{Diego Andre Eifer}
\ead{deifer@hcpa.edu.br}

\affiliation[inst2]{department={Radiology Service}, organization={Hospital de Clínicas de Porto Alegre (HCPA)},
            addressline={Ramiro Barcelos, 2350}, 
            city={Porto Alegre},
            postcode={90035-903}, 
            State={Rio Grande do Sul},
            country={Brazil}
            }
            
\author[inst3]{Mauricio Anes}
\ead{manes@hcpa.edu.br}

\author[inst3]{Guilherme Ribeiro Garcia}
\ead{grgarcia@hcpa.edu.br}

\affiliation[inst3]{department={Medical Physics Service}, organization={Hospital de Clínicas de Porto Alegre (HCPA)},
            addressline={Ramiro Barcelos, 2350}, 
            city={Porto Alegre},
            postcode={90035-903}, 
            state={Rio Grande do Sul},
            country={Brazil}
            }

\affiliation[inst4]{department={Bioinformatics Core},Organization={Hospital de Clínicas de Porto Alegre (HCPA)},
            addressline={Ramiro Barcelos, 2350}, 
            city={Porto Alegre},
            postcode={90035-903}, 
            state={Rio Grande do Sul},
            country={Brazil}
            }
        
\author[inst1,inst4]{Mariana Recamonde-Mendoza}
\ead{mrmendoza@inf.ufrgs.br}

\begin{abstract}
Artificial intelligence (AI) is increasingly being utilized to optimize magnetic resonance imaging (MRI) protocols. Given that image details are critical for diagnostic accuracy, optimizing MRI acquisition protocols is essential for enhancing image quality. While medical physicists are responsible for this optimization, the variability in equipment usage and the wide range of MRI protocols in clinical settings pose significant challenges. This study aims to validate the application of AI in optimizing MRI protocols using dynamic data from clinical practice, specifically DICOM metadata. To achieve this, four MRI spine exam databases were created, with the target attribute being the binary classification of image quality (good or bad). Five AI models were trained to identify trends in acquisition parameters that influence image quality, grounded in MRI theory. These trends were analyzed using SHAP graphs. 
The models achieved F1 performance ranging from 77\% to 93\% for datasets containing 292 or more instances, with the observed trends aligning with MRI theory. The models effectively reflected the practical realities of clinical MRI settings, offering a valuable tool for medical physicists in quality control tasks. In conclusion, AI has demonstrated its potential to optimize MRI protocols, supporting medical physicists in improving image quality and enhancing the efficiency of quality control in clinical practice.
\end{abstract}

\begin{keyword}
{Artificial Intelligence \sep Magnetic Resonance Imaging \sep Acquisition Protocol \sep Optimization \sep Machine Learning}
\end{keyword}

\end{frontmatter}


\section{Introduction}
\label{sec:intro}
Radiology is essential in medicine, standing out for its pivotal role in diagnostics and treatment monitoring. Legislation in many countries is in force to ensure the quality of radiological services \cite{rdc, europeanRegulation, canadianRegulation, japaneseRegulation}. Typically, medical physicists are tasked with implementing quality assurance programs in radiology, as outlined by international guidelines and agencies \cite{AAPM, AFOMP, EFOMP, IAEA}.

An optimal quality of the images produced in magnetic resonance imaging (MRI) equipment is crucial, as subtle details can be critical for accurate and timely diagnoses. Therefore, one of the guidelines of the quality program is the commitment to the continuous optimization of the protocols of the image acquisition sequences. Such optimization seeks to improve the quality of the images, and reduce costs, errors, and the need for repeat exams. This practice involves fine-tuning acquisition parameters to account for specific system characteristics and real-world operating conditions.  MRI technologists, as highlighted in the literature \cite{cat1, cat2}, are encouraged to make these adjustments under the guidance and monitoring of medical physicists through continuing education programs.

However, monitoring these adjustments presents significant challenges, such as the variability in technologist-configured parameters, the complexity of balancing image quality with acquisition time, and the limited resources available in hospital settings. Recent advancements suggest that artificial intelligence (AI) could address these challenges by optimizing MRI protocols \cite{optimStudy, mrZero, samplingMri}. For instance, efforts such as the development of AI-driven sampling techniques to accelerate image reconstruction and machine learning (ML) models to fine-tune sequence parameters have shown promise. Nevertheless, many studies still rely on idealized simulation scenarios rather than real-world clinical data, highlighting the need for further research in practical applications.

In this context, this work proposes the development and evaluation of an AI-based methodology to assist in the optimization of MRI protocols from real and dynamic data, using the DICOM\footnote{Digital Imaging and Communications in Medicine.} image metadata information extracted from spine MRI sequences. The results demonstrated that for datasets with 292 instances or more, the ML models identified trends consistent with MRI theory, achieving F1-scores between 77\% and 93\% in the task of predicting good quality images.

The remainder of this paper is organized as follows: Section~\ref{sec:background} provides the theoretical background for validating the results; Section~\ref{sec:literature} reviews related studies; Section~\ref{sec:method} details our methodology, from data collection and preparation to AI model development; Section~\ref{sec:results} presents the findings; Section~\ref{sec:discussion} evaluates the results in light of MRI theory; and Section~\ref{sec:conclusion} summarizes the contributions and implications of our study.

\section{Background} 
\label{sec:background}
This section briefly presents the theory underlying parameter adjustments and their effects on spinal image quality. The theoretical discussion of MRI image quality will focus on two key aspects: Signal-to-Noise Ratio (SNR) and spatial resolution. In MRI, the SNR represents the ratio between the signal from structures of interest and the intrinsic background noise. A higher SNR is desirable, as it ensures better definition of anatomical structures while minimizing noise interference. Spatial resolution, on the other hand, refers to the ability to distinguish small adjacent structures in an image. The smaller the structure that can be discerned from its surroundings, the better the spatial resolution \cite{acr, cat1}.

Regarding the parameters, a comprehensive review is beyond the scope of this work but can be found in standard MRI textbooks \cite{cat1, handbookPhysics,  parametersBkg}. Here, we address the parameters of interest, which are those adjustable between MRI acquisitions and that have a well-described impact on image quality in the literature. 

The repetition time (RT) refers to the period, measured in milliseconds, between the radiofrequency pulses used to orient the magnetization\footnote{The magnetization is generated by the effect of the MRI machine's magnetic field on the magnetic spins of atomic nuclei, such as hydrogen} in the detection plane of the equipment. The resulting vector of this magnetization has its magnitude altered depending on changes in the duration of this period. With a short RT, the magnetization resulting along the detection axis is reduced, leading to a lower SNR in the resulting image. However, a short TR is crucial for achieving the desired contrast in T1-weighted images, which depend on the behavior of the magnetization's return to its original axis. In contrast, increasing the TR generally results in a higher SNR in the image, although this also prolongs the scan time.

The Field of View (FOV) represents the size of the image acquisition area. In spinal MRI, the primary objective is often to enhance image resolution, which can be achieved by reducing the FOV. However, excessive FOV reduction, even when aimed at focusing on spinal structures, may result in highly noisy images, potentially compromising diagnostic accuracy. Additionally, reducing the FOV results in approximately a 40\% loss of SNR while increasing spatial resolution by only about 20\% \cite{cat1, parametersBkg}. Therefore, unless maximizing spatial resolution is the primary objective, increasing the FOV generally enhances the overall amount of acquired information, which theoretically leads to improved image quality despite a slight compromise in resolution \cite{cat1, parametersBkg}.

The Percent Phase Field of View (pFOV) represents the percentage of the field size in the phase-encoding direction relative to the frequency-encoding direction. In spinal MRI, reducing the FOV in the phase-encoding direction is often employed to decrease scan time. However, this reduction negatively impacts the SNR. If scan time is not a limiting factor, increasing the pFOV enhances the amount of acquired information and, consequently, image quality. Notably, the SNR is proportional to the square root of the number of phase-encoding steps  \cite{cat1, parametersBkg}.

The Number of Excitations (NEX) represents the number of times a k-space line is sampled, while percent sampling indicates the percentage of total k-space lines used to reconstruct the image. Both parameters significantly impact scan duration, which is particularly relevant in spinal exams due to patient discomfort during prolonged acquisitions \cite{nexSpine}. Reducing a high NEX may compromise both SNR and spatial resolution, but it also reduces motion artifacts and subsequent image blurring. Additionally, the relationship between NEX and SNR is expressed as $SNR \propto \sqrt{NEX}$, indicating that substantial increases in NEX are required to meaningfully improve SNR \cite{parametersBkg}.

Slice thickness determines the thickness of each 2D image slice. Ideally, slices should be as thin as possible to minimize structural overlap and avoid signal averaging within a pixel. However, excessive reductions in slice thickness can significantly reduce SNR, particularly in acquisitions where the overall signal is inherently low. Furthermore, thinner slices increase scan time \cite{parametersBkg}.

\section{Related Work}
\label{sec:literature}

This literature review examines the current advancements in applying AI to optimize medical protocols and explores the integration of DICOM metadata as attributes in database construction, emphasizing its potential in enhancing MRI protocol optimization. By leveraging ML to analyze DICOM-derived image information, these methodologies highlight the versatility of AI in streamlining workflows, improving diagnostic accuracy, and adapting protocols to clinical needs. Notably, no prior studies were identified that directly align with the methodology proposed in this work, underscoring its originality and potential contributions to the field.

\subsection{Optimization of Medical Protocols Using Artificial Intelligence} \label{AIinOptMedical}

AI, particularly through ML methodologies, has demonstrated remarkable potential in optimizing medical protocols and workflows. By identifying key attributes that significantly influence target outcomes, ML-based classification methods facilitate the generation of optimized responses tailored to specific clinical processes \cite{reviewOptMedical}.

A notable application of AI in healthcare optimization is the customization of treatment plans. By integrating diverse data sources such as clinical attributes, lifestyle factors, and environmental conditions, AI models can recommend personalized therapies. For instance, in chronic diabetes management, neural network architectures like Long Short-Term Memory (LSTM), evaluated using holdout and five-fold cross-validation methods, have shown promising results \cite{diabetes}.

AI's role extends to optimizing scheduling and resource allocation in healthcare facilities. Gradient Boosting (GB) and Decision Tree (DT) models, for example, have successfully reduced no-show rates by up to 19\% through predictive reminders based on scheduling attributes. These models underwent rigorous validation using nested cross-validation and an 80/20 train-test split \cite{noShows}. Similarly, Random Forest (RF) and Support Vector Machine (SVM) algorithms have been applied to surgical scheduling, improving resource utilization while minimizing patient wait times \cite{reviewOptMedical, optSchedule}. Moreover, predictive analysis powered by AI has been used to enhance resource preparation, leveraging historical patient admission, transfer, and discharge data to identify seasonal trends and better allocate resources \cite{reviewOptMedical}.

In the fields of medical physics and bioengineering, AI is increasingly utilized for predictive maintenance of medical equipment. Models trained on data from Computerized Maintenance Management Systems—such as Decision Trees (DT), K-Nearest Neighbors (KNN), Naïve Bayes (NB), Support Vector Machines (SVM), Random Forest (RF), and Artificial Neural Networks (ANN)—have shown significant promise. Validated through ten-fold cross-validation, these models have benefited from Bayesian optimization for hyperparameter tuning and achieved good performance, thereby improving the efficiency of maintenance scheduling \cite{optMaint}.

Overall, AI-based protocol optimization typically relies on interpretable ML models, particularly when analyzing structured datasets to identify critical attributes. This focus on interpretability not only ensures reliability but also enhances the practical applicability of these models in clinical settings.

\subsection{Usage of DICOM Image Metadata in AI Studies}
The increasing adoption of AI methodologies in medical imaging has been facilitated by advancements in extracting and analyzing metadata from DICOM datasets \cite{dicomExtraction}. In computed tomography imaging, AI models have been trained using DICOM metadata to optimize contrast detection and identify intravenous contrast phases. Convolutional Neural Networks (CNNs), optimized through nested cross-validation with an 80/10/10 data split (training, validation, test), have achieved accuracies exceeding 93\% \cite{dicomCT}. 

In mammography, GB and Deep Learning (DL) models have been applied to predict breast tissue deformation during compression. Attributes such as compression thickness, extracted from DICOM metadata, were critical inputs. These models achieved root mean square error values ranging from 0.47 mm to 1.70 mm \cite{dicomMammo}. DICOM data has also been used in radiological forensic imaging to optimize preprocessing steps. Metadata aids in separating and categorizing image volumes for specific processing tasks, whether in 2D or reconstructed 3D formats \cite{dicomForensics}. 

Another study applied AI to assist technologists in selecting optimal positioning and detection regions in spinal MRI. Region-based CNN models were trained and validated using a 64/26/10 data split, with hyperparameters fine-tuned through cross-validation. Approximately 55\% of suggested parameters required minor adjustments in practical use \cite{dicomScanPrescription}. Additionally, DICOM metadata has been leveraged to classify MRI acquisition protocols using Random Forest (RF) models, achieving an average precision of 86\% and an F1-score of 84\% across 16 protocol classes \cite{dicomClassifier}.

These findings underscore the versatility and relevance of DICOM metadata in AI-based medical imaging applications, providing robust datasets for diverse ML methodologies.

\subsection{Application of AI in Optimizing Magnetic Resonance Imaging Protocols}
Recent research has increasingly focused on optimizing MRI protocols to reduce scan times while maintaining diagnostic image quality. DL models have played a pivotal role in this area by reconstructing high-quality images from undersampled k-space data, effectively shortening acquisition times without compromising diagnostic value \cite{samplingMri,acceleratedMri,mriHighQuality,undersampleRec}. Further advancements in contrast techniques have utilized DL to generate optimized MRI sequences from scratch. These approaches often rely on data from simulated scanner environments to build and validate sequences both in simulated settings and \textit{in vivo} \cite{mrZero}.

In addition to sequence optimization, parameter tuning has emerged as a critical focus in MRI protocol studies. ML models such as Support Vector Regression (SVR), KNN, and RF, researchers have optimized imaging parameters based on data derived from simulated scanner environments. Standard practices, such as splitting datasets into 60/20/20 (training, validation, test) and employing ten-fold cross-validation, were used for hyperparameter tuning and model evaluation. Among the tested models, SVR and KNN performed the best, with root mean square error (RMSE) values ranging from 0.00 to 11.50 for SVR and 0.04 to 22.30 for KNN. Despite these promising results, the reliance on simulated data (phantoms) underscores the need for methodologies validated on real-world clinical datasets \cite{optimStudy}.

A complementary avenue of research has focused on assessing MRI image quality using AI. One study developed a binary classification convolutional neural network (CNN) to categorize MRI images as either "good" or "poor" quality. The training dataset incorporated DICOM images sourced from multiple MRI devices and public repositories, such as The Cancer Imaging Archive (TCIA) \cite{tcia}. Additional attributes, including scan duration, body part, and signal-to-noise ratio (SNR), were extracted from DICOM metadata and included in the dataset. Ground truth labels were determined by expert radiologists based on practical quality indicators, such as motion artifacts, signal non-uniformity, and magnetic susceptibility effects. The model achieved an F1-score of 89.2\% on the validation set and 90.0\% on the test set. However, the test specificity was notably low at 22.4\%, highlighting areas for improvement \cite{imageQualityClass}.

Collectively, these studies emphasize the growing integration of AI into MRI protocol optimization. The results demonstrate its potential to streamline image acquisition workflows, enhance diagnostic accuracy, and adapt protocols to meet specific clinical needs, paving the way for more efficient and patient-centered imaging practices.

\section{Material and Methods} 
This section details the processes of data collection and pre-processing, model training and evaluation, and the experimental methodologies adopted in this study. 

\label{sec:method}
\subsection{Data Collection and Pre-Processing} 
\label{sub:database}
For MRI protocol optimization, it was essential to gather data from a single protocol, recorded on the same equipment within the same healthcare facility, ensuring consistency and minimizing hardware-related biases. While public DICOM databases (\eg \cite{tcia,gbmDataset}) were considered, limitations in protocol and equipment consistency led to the construction of a custom dataset in collaboration with the Radiology Service and the Medical Physics Service at Hospital de Clínicas de Porto Alegre (HCPA).

Using the Enterprise Viewer 8.12, a database query was conducted to retrieve MRI examinations based on the following criteria: Philips Achieva 1.5T equipment, cervical and lumbosacral spine imaging, patients aged 18 years or older, and examinations performed between January 1, 2016, and October 31, 2023. Incomplete examinations were excluded, and all data were anonymized using DicomCleaner™. The final dataset comprised 668 lumbosacral and 679 cervical spine MRI examinations. This study was reviewed and approved by the Research Ethics Committee of Hospital de Clínicas de Porto Alegre (CAAE number 74933423.2.0000.5327).

A Python script was developed to automate slice selection, categorizing images based on examination type, acquisition protocol, coil, and acquisition plane. The largest subsets identified were sagittal T1 (ST1) and T2 (ST2) protocols for both cervical and lumbosacral spine regions, yielding datasets comprising 292 samples for lumbosacral spine (LS) ST1, 237 for LS ST2, 374 for cervical spine (C) ST1, and 357 for C ST2.

To validate the application of AI in MRI protocol optimization, image quality metrics were calculated using entropy power~\footnote{Entropy power quantifies the uncertainty or variability of a signal's distribution. In the context of images, higher values may indicate greater randomness or complexity, which could correspond to noisier patterns.} and spectral flatness~\footnote{Spectral flatness is a frequency-domain measure that assesses the uniformity of a signal's power spectrum. Higher values indicate that the signal's spectrum resembles white noise, potentially corresponding to less structured regions in an image.}, as discussed in \citet{expertsQualityPerception}. A second Python script extracted DICOM attributes and computed these metrics, producing a structured dataset for analysis. Finally, the target variable was defined by normalizing entropy power and spectral flatness values. Image quality was categorized relative to the median, with lower target values corresponding to higher-quality images (class 1) and higher values indicating lower-quality images (class 0).

DICOM attributes were divided into two groups: commonly modified parameters (\eg slice thickness, repetition time [RT], echo time [TE], and field of view [FOV]) and randomly modified parameters (\eg age, weight, and gender). Dimensionality reduction was performed using Pearson and Spearman correlation analyses, eliminating attributes with correlation coefficients exceeding 0.7 in both methods or 0.9 in one method.

At the end of the pre-processing step, four datasets -- LS-ST1, LS-ST2, C-ST1, and C-ST2 -- were prepared for subsequent ML analyses.

\subsection{ML Models Development} 
\label{sub:models}

Our study employed a range of binary classification algorithms commonly used in the literature, selected for their effectiveness and interpretability in overall and related ML tasks. The chosen algorithms included Logistic Regression (LR), Decision Trees (DT), Random Forest (RF), and Gradient Boosting (GB). Additionally, a simple deep learning model, the Multilayer Perceptron (MLP), was incorporated to assess the potential benefits of deep learning techniques for this application.

To ensure consistency and comparability, a standardized methodology was applied to all four datasets. Pre-processing steps included attribute normalization, which was specifically necessary for LR and MLP models due to their sensitivity to feature scaling. Each dataset was split using the hold-out method, allocating 80\% for training and 20\% for testing.

Model training and hyperparameter optimization were conducted on the Google Colab platform, leveraging a T4 GPU to enhance computational efficiency. A nested cross-validation (NCV) approach was employed during training to ensure robust performance evaluation and optimal hyperparameter tuning. Grid search was used within the inner NCV loop (three folds) to identify the best hyperparameter configurations, with the F1-score serving as the primary optimization metric (see Section~\ref{sub:evaluation} for further details on model evaluation).

The outer NCV loop (ten folds) provided an unbiased assessment of model performance, ensuring that the selected hyperparameters generalized well to unseen data. This rigorous validation strategy minimized the risk of overfitting and provided reliable performance estimates for each algorithm. By following this structured approach, the study ensured that all models were trained and evaluated consistently, enabling a robust comparison of their effectiveness in addressing the proposed classification task.

\subsection{Performance Evaluation} 
\label{sub:evaluation}
The F1-score was selected as the primary metric for performance evaluation during model training and evaluation due to its ability to balance precision and recall, making it particularly suited for evaluating performance in scenarios where sensitivity to both false positives and false negatives is critical. Additional metrics, including accuracy, precision, recall, the area under the ROC curve (AUC-ROC), and the area under the precision-recall curve (AUC-PR) were also calculated to provide a comprehensive view of model performance, especially for imbalanced datasets. Mean values and standard deviations were reported for all metrics to assess consistency and robustness.

The final model was constructed using the set of hyperparameters that either consistently appeared as the optimal configuration or achieved the highest overall performance during the nested cross-validation phase. This approach ensured that the chosen model represented the best balance between predictive accuracy and generalization capacity. To assess potential overfitting and validate the generalization of the model to unseen data, the finalized model was evaluated on the 20\% hold-out test set. 

\subsection{Model Interpretability}

Model explainability was addressed using SHAP (SHapley Additive exPlanations) values, which were visualized using beeswarm plots to highlight how variations in attribute values influenced the model's classification of image quality \cite{shapOriginal,MlInterpretable}. To enhance performance for the tree-based models, DF, RF, and GB, the SHAP Tree Explainer was used, and for MLP and LR, the SHAP Kernel Explainer was used.

To facilitate a broader understanding of attribute importance across models, a summary visualization was developed for each dataset. This visualization integrated qualitative insights from SHAP importance rankings with the quantitative performance of the models, as measured by their F1-scores on the test data. Attribute intersections across models were depicted using color gradients, where the intensity reflected both the SHAP importance ranking and the respective model's performance, ensuring a balanced representation of these factors. This iterative analysis was conducted separately for each dataset, ensuring that trends specific to different protocols or regions of the spine were adequately captured.

\section{Results}
\label{sec:results}

This section presents the results of the proposed methodology, focusing on the performance evaluation of the models and the analysis of feature impacts on image quality predictions. The results are systematically reported for the datasets C-ST1, LS-ST1, C-ST2, and LS-ST2, in the order presented.

\subsection{C-ST1 Dataset}
\label{sub:cst1}

The results of the performance evaluation for the C-ST1 dataset, conducted based on the nested CV process, are summarized in Table~\ref{tab:CS1performance}. Boxplot graphs are available in the Supplementary Material, Figure~1. The model that demonstrated the best performance, based on the mean F1-score, was GB, achieving a value of 0.83 ± 0.08. This model exhibited a good balance between precision (0.86 ± 0.09) and recall (0.81 ± 0.11), and also performed well in terms of accuracy (0.83 ± 0.06), further reinforcing its overall performance. Ensemble-based models, GB and RF, stood out in terms of AUC-PR. However, RF showed a poorer recall (0.78 ± 0.14), along with DT (0.78 ± 0.12). Finally, MLP yielded more modest outcomes in terms of accuracy (0.80 ± 0.06) and precision (0.80 ± 0.07), achieving F1-scores of 0.80 ± 0.06.

The optimal hyperparameters for each model trained on the C-ST1 dataset (available in Table~1 of the Supplementary Material) were determined during the training phase and subsequently applied to train the final models for each algorithm. These final models were evaluated on a hold-out test set, with the results summarized in the bottom rows of Table~\ref{tab:CS1performance}. Overall, the test set results closely mirrored the performance distribution observed during the nested cross-validation process, with most metrics falling within the interquartile range of the validation scores (Supplementary Material, Figure~1). This consistency highlights the robustness of the trained models. GB and RF exhibited strong performance across all metrics, excelling in accuracy, precision, recall, and F1-score, demonstrating their effectiveness in distinguishing between quality classes. Conversely, LR and DT displayed limited generalization capacity. Notably, RF and the MLP achieved the highest AUC-PR values on the test dataset, exceeding expectations based on their validation performance.

\begin{table}[t]

\resizebox{\textwidth}{!}{%
\begin{tabular}{llcccccc}
\textbf{Model} & \textbf{Stage} &\textbf{Accuracy} & \textbf{Precision} & \textbf{Recall} & \textbf{AUC-PR} & \textbf{AUC-ROC} & \textbf{F1} \\ \cline{1-8} 
\textbf{LR}  &  \multirow{5}{*}{NCV}   & 0.81 ± 0.06 & 0.83 ± 0.09 &  0.80 ± 0.11 & 0.87 ± 0.06 & 0.85 ± 0.06 & 0.81 ± 0.07 \\
\textbf{DT}  &  & 0.79 ± 0.06 & 0.80 ± 0.06 & 0.78 ± 0.12 & 0.76 ± 0.07 & 0.81 ± 0.06 & 0.78 ± 0.07 \\
\textbf{RF}  &   &  0.80 ± 0.06 & 0.83 ± 0.09 & 0.78 ± 0.14 & 0.91 ± 0.04 & 0.89 ± 0.05 & 0.79 ± 0.08 \\
\textbf{GB}  &      & 0.83 ± 0.06 & 0.86 ± 0.09 & 0.81 ± 0.11 & 0.91 ± 0.05 & 0.90 ± 0.06 & 0.83 ± 0.08 \\
\textbf{MLP} & & 0.80 ± 0.06 & 0.80 ± 0.07 & 0.81 ± 0.09 & 0.86 ± 0.05 & 0.86 ± 0.06 & 0.80 ± 0.06 \\  \hline
\textbf{LR}   &   \multirow{5}{*}{Test}    & 0.80 & 0.76 & 0.86 & 0.90 & 0.87 & 0.81\\
\textbf{DT}  &  & 0.80 & 0.78 & 0.84 & 0.82 & 0.85 & 0.81 \\
\textbf{RF}  &   & 0.84 & 0.80 & 0.89 & 0.91 & 0.91 & 0.85 \\
\textbf{GB}  &         & 0.84 & 0.80 & 0.89 & 0.89 & 0.90 & 0.85 \\
\textbf{MLP} & & 0.83 & 0.79 & 0.89 & 0.92 & 0.91 & 0.84 \\ \hline
\end{tabular}
}
\caption{Performance metrics of the models trained on the C-ST1 dataset, evaluated using nested cross-validation (NCV) and test data. \label{tab:CS1performance}}
\end{table}

\begin{figure}[!h]
     \centering
     \begin{subfigure}[b]{0.47\textwidth}
         \centering
         \includegraphics[width=\textwidth]{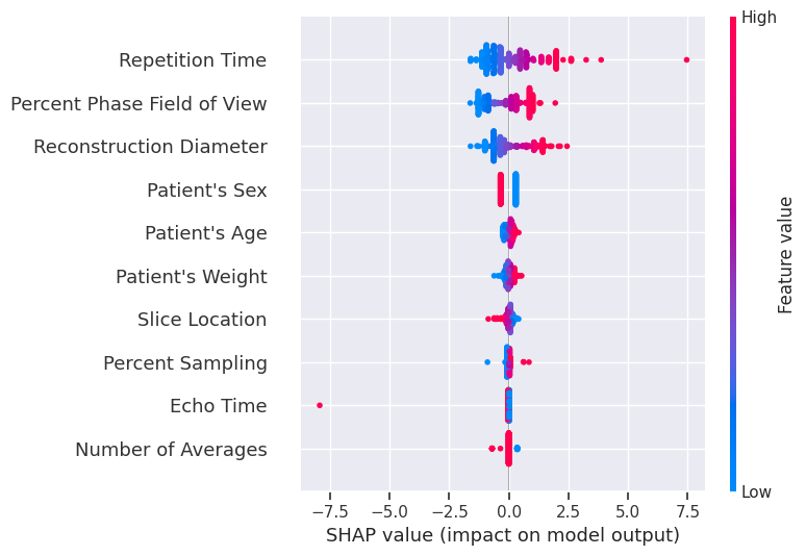}
         \caption{LR}
         \label{fig:CS1shap:LR}
     \end{subfigure}
     \hfill
     \begin{subfigure}[b]{0.47\textwidth}
         \centering
         \includegraphics[width=\textwidth]{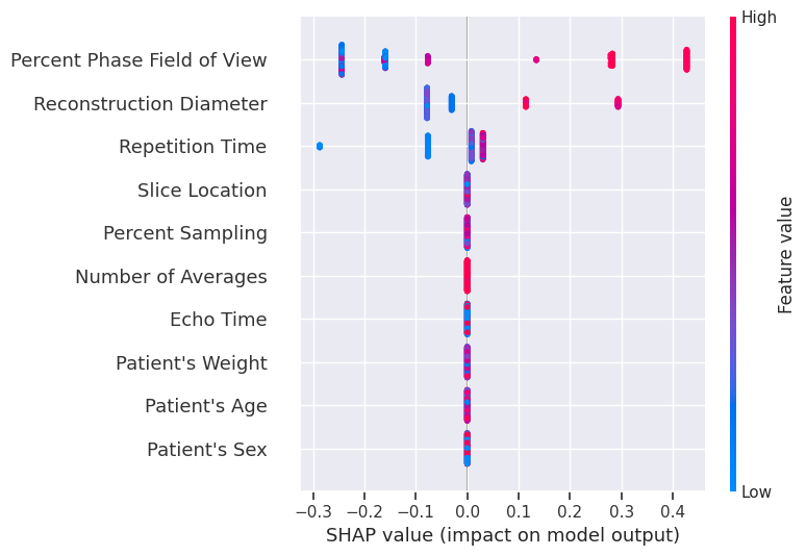}
         \caption{DT}
         \label{fig:CS1shap:DT}        
     \end{subfigure}
     \\
     \begin{subfigure}[b]{0.47\textwidth}
         \centering
         \includegraphics[width=\textwidth]{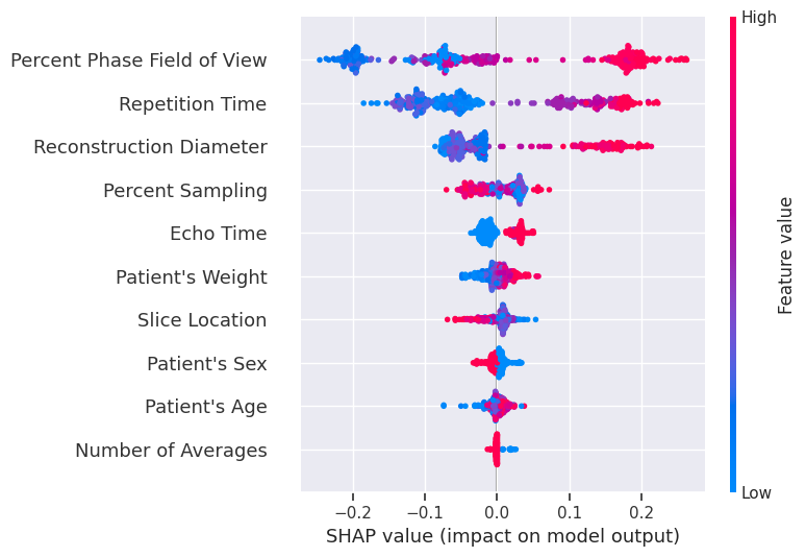}
         \caption{RF}
         \label{fig:CS1shap:RF}
     \end{subfigure}
     \hfill
     \begin{subfigure}[b]{0.47\textwidth}
         \centering
         \includegraphics[width=\textwidth]{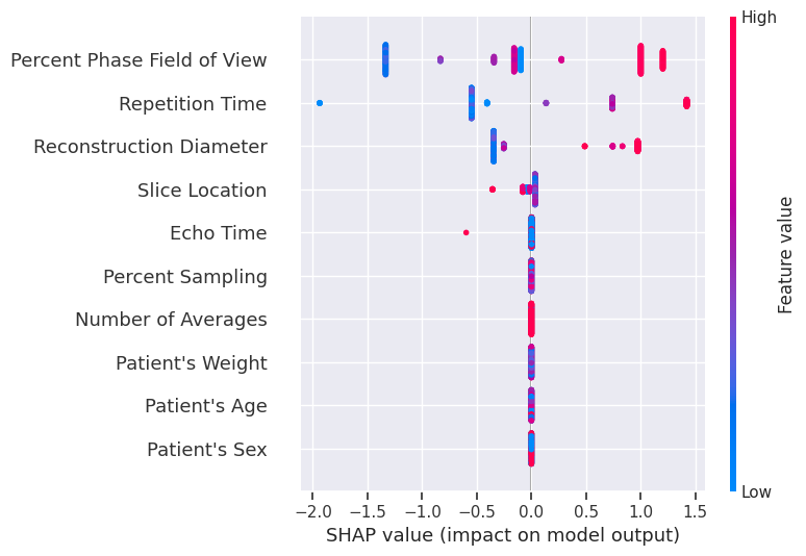}
         \caption{GB}
         \label{fig:CS1shap:GB}        
     \end{subfigure}
      \\
     \begin{subfigure}[b]{0.47\textwidth}
         \centering
         \includegraphics[width=\textwidth]{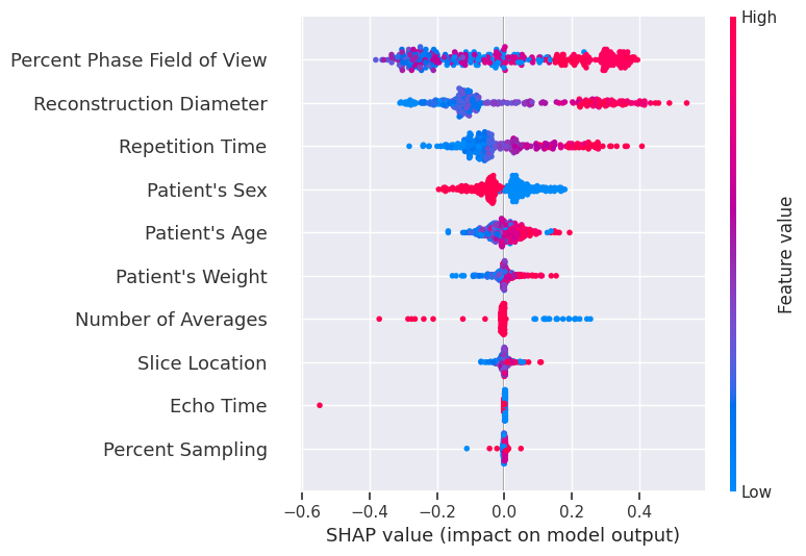}
         \caption{MLP}
         \label{fig:CS1shap:MLP}
     \end{subfigure}
     \hfill
     
\caption{SHAP values results for the features in the C-ST1 dataset.}
\label{fig:CS1shap}
\end{figure}

To evaluate the impact of features on image quality classification, we conducted a SHAP analysis, with results presented in Figure~\ref{fig:CS1shap}. It is important to note that the Philips Achieva 1.5T equipment considers the reconstruction diameter as the largest dimension of the FOV, which is treated as equivalent to the FOV in this study \cite{philipsDicom}. From the SHAP analysis, we observed that in top-performing models (GB and RF), higher values of pFOV, FOV, and RT are associated with better image quality. However, insufficient data across all models prevented confirming trends for TE and Sampling Percentage. Additionally, in some models, the NEX showed a slight trend towards improved image quality when its value is lower. 

\begin{figure}[!th]
\centering
\includegraphics[width=0.8\textwidth]{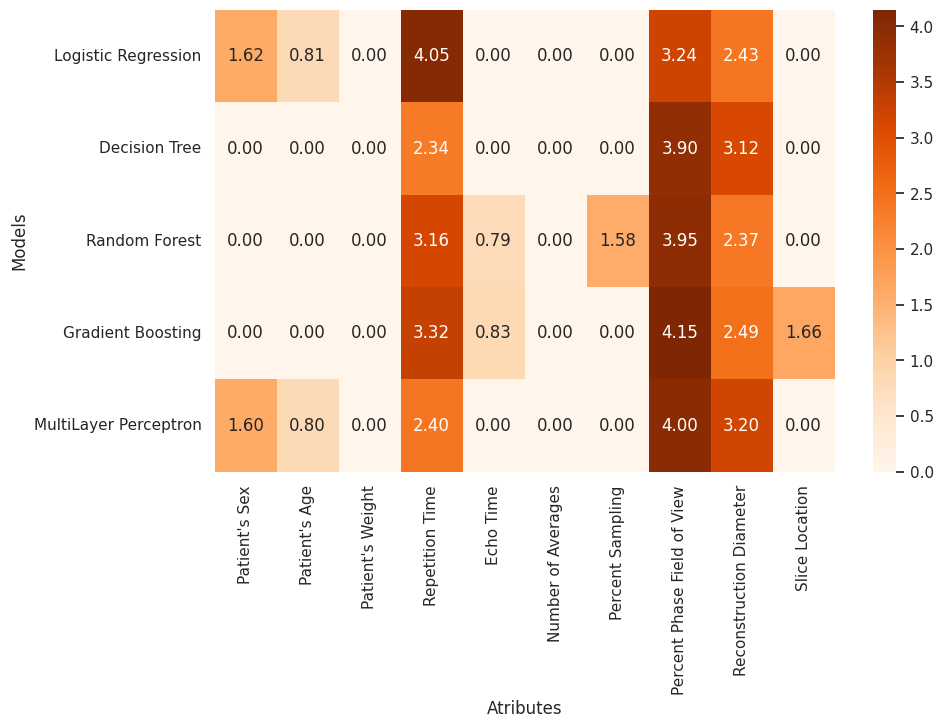}
\caption{Top five most important features for each model, weighted by their respective F1-Score results for the C-ST1 dataset.}
\label{fig:CS15models}
\end{figure}

Figure~\ref{fig:CS15models} highlights the most significant features influencing image quality in cervical spine acquisitions from the ST1 protocol. In summary, the three most significant features affecting image quality in the C-ST1 dataset are pFOV, FOV, and TR, as indicated by the SHAP analysis.

\subsection{LS-ST1 Dataset}
\label{sub:lsst1}

The performance assessment for the LS-ST1 dataset is presented in Table~\ref{tab:LS1performance}. Detailed visualizations of metric distributions from the NCV process are provided in the Supplementary Material, Figure~2. Analyzing the NCV scores, RF achieved the highest F1-score (0.82 ± 0.06) and AUC-PR (0.90 ± 0.07), along with consistent precision (0.80 ± 0.07) and recall (0.85 ± 0.11), demonstrating its strong ability to distinguish between classes. GB followed closely, with an F1-score of 0.81 ± 0.06 and well-balanced recall and precision values. While LR, MLP, and DT displayed competitive recall, their lower precision led to a decline in overall performance for this dataset.

On the hold-out test set, the final models trained with optimized hyperparameters, available in Table~2 of the Supplementary Material, displayed performance trends consistent with those observed during the NCV process. GB and DT achieved the highest F1-scores (0.87), demonstrating their ability to effectively balance recall and precision on unseen data. RF stood out with the highest AUC-PR value (0.94), reflecting its superior ability to accurately classify positive instances across varying thresholds. LR showed a marked improvement in recall (0.93), reducing the performance gap compared to the ensemble models. While the MLP achieved recall comparable to the top-performing models, its lower precision (0.75) adversely impacted its F1-score. Overall, these results emphasize RF's robustness and consistent performance across metrics, being selected as the top-performing model for this dataset, while also highlighting GB's adaptability to unseen data.

\begin{table}[t]
\resizebox{\textwidth}{!}{%
\begin{tabular}{llcccccc}
\textbf{Model} & \textbf{Stage} &\textbf{Accuracy} & \textbf{Precision} & \textbf{Recall} & \textbf{AUC-PR} & \textbf{AUC-ROC} & \textbf{F1} \\ \cline{1-8} 
\textbf{LR}  &  \multirow{5}{*}{NCV}     & 0.77 ± 0.07 & 0.75 ± 0.08 & 0.85 ± 0.12 & 0.82 ± 0.10 & 0.86 ± 0.06 & 0.79 ± 0.07 \\
\textbf{DT}  &  & 0.77 ± 0.06 & 0.78 ± 0.09 & 0.80 ± 0.14 & 0.78 ± 0.09 & 0.82 ± 0.08 & 0.77 ± 0.07 \\
\textbf{RF}  &   & 0.81 ± 0.06 & 0.80 ± 0.07 & 0.85 ± 0.11 & 0.90 ± 0.07 & 0.90 ± 0.06 & 0.82 ± 0.06 \\
\textbf{GB}  &        & 0.80 ± 0.06 & 0.81 ± 0.08 & 0.81 ± 0.09 & 0.80 ± 0.08 & 0.85 ± 0.06 & 0.81 ± 0.06 \\
\textbf{MLP} & & 0.77 ± 0.07 & 0.75 ± 0.09 & 0.84 ± 0.13 & 0.82 ± 0.09 & 0.86 ± 0.07 & 0.78 ± 0.07 \\ \hline
\textbf{LR} &  \multirow{5}{*}{Test}    & 0.85 & 0.79 & 0.93 & 0.87 & 0.90 & 0.86 \\
\textbf{DT} &        & 0.86 & 0.82 & 0.93 & 0.80 & 0.87 & 0.87 \\
\textbf{RF} &        & 0.85 & 0.79 & 0.93 & 0.94 & 0.93 & 0.86 \\
\textbf{GB} &      & 0.86 & 0.82 & 0.93 & 0.86 & 0.91 & 0.87 \\
\textbf{MLP} & & 0.81 & 0.75 & 0.93 & 0.86 & 0.89 & 0.83 \\ \hline
\end{tabular}
}
\caption{Performance metrics of the models trained on the LS-ST1 dataset, evaluated using nested cross-validation (NCV) and test data.\label{tab:LS1performance}}

\end{table}

\begin{figure}[!h]
     \centering
     \begin{subfigure}[b]{0.47\textwidth}
         \centering
         \includegraphics[width=\textwidth]{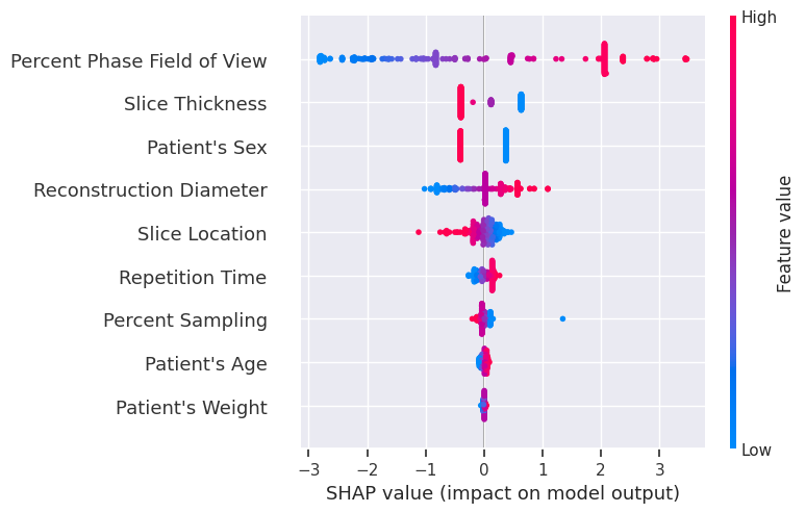}
         \caption{LR}
         \label{fig:LS1shap:LR}
     \end{subfigure}
     \hfill
     \begin{subfigure}[b]{0.47\textwidth}
         \centering
         \includegraphics[width=\textwidth]{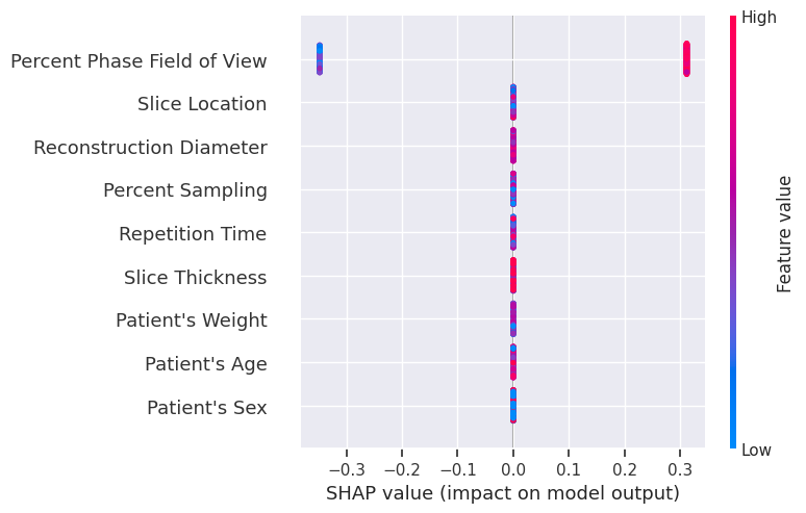}
         \caption{DT}
         \label{fig:LS1shap:DT}        
     \end{subfigure}
     \\
     \begin{subfigure}[b]{0.47\textwidth}
         \centering
         \includegraphics[width=\textwidth]{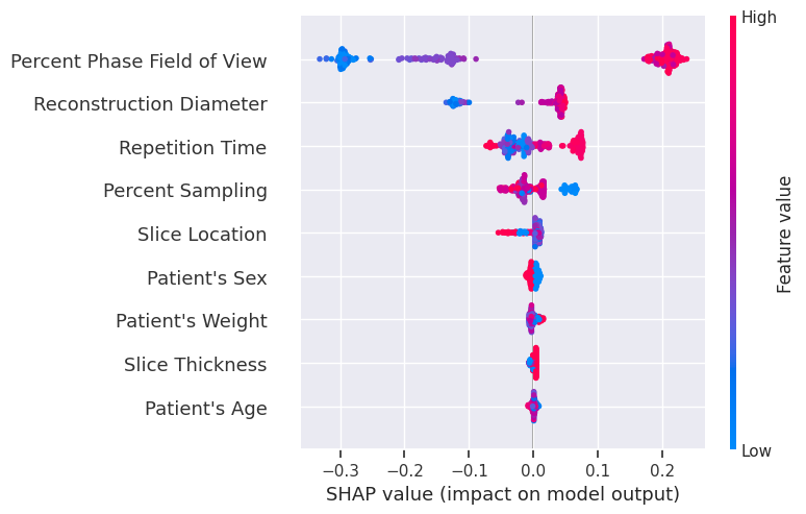}
         \caption{RF}
         \label{fig:LS1shap:RF}
     \end{subfigure}
     \hfill
     \begin{subfigure}[b]{0.47\textwidth}
         \centering
         \includegraphics[width=\textwidth]{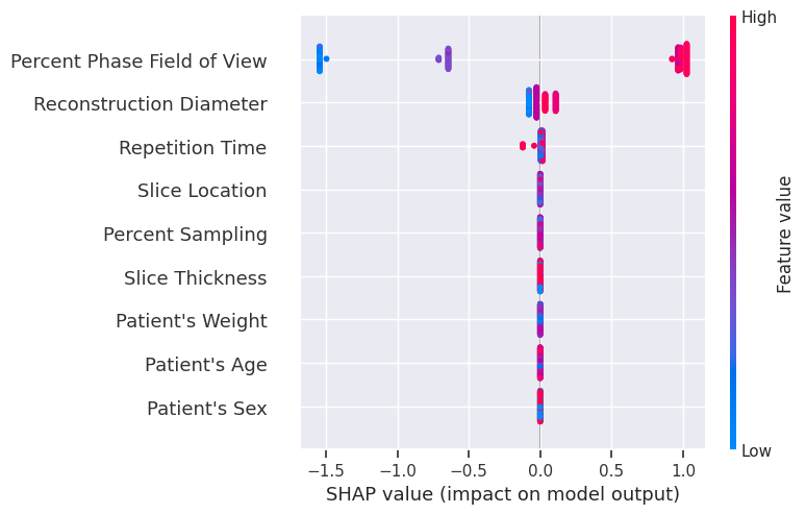}
         \caption{GB}
         \label{fig:LS1shap:GB}        
     \end{subfigure}
      \\
     \begin{subfigure}[b]{0.47\textwidth}
         \centering
         \includegraphics[width=\textwidth]{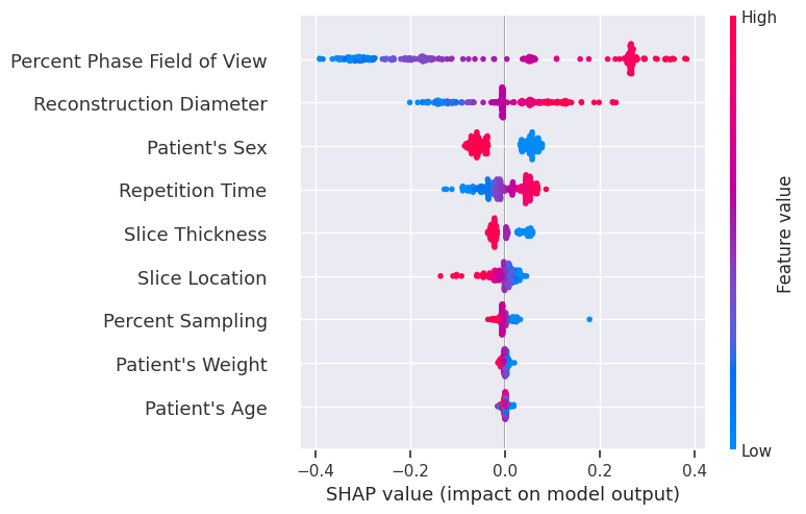}
         \caption{MLP}
         \label{fig:LS1shap:MLP}
     \end{subfigure}
     \hfill
\caption{SHAP values results for the features in the LS-ST1 dataset.}
\label{fig:LS1shap}
\end{figure}

The SHAP analysis presented in Figure~\ref{fig:LS1shap} assesses the influence of acquisition parameters on image quality in this protocol. For RF, the best-performing model in this evaluation, a clear trend was observed: higher values of pFOV, FOV, and RT were associated with superior image quality. Additionally, pFOV and FOV consistently ranked among the top four most influential features across all models, emphasizing their importance in determining image quality. A notable pattern also emerged, indicating that a combination of lower sampling percentage and slice thickness, along with higher pFOV, RT, and FOV values, contributes to improved image quality.

\begin{figure}
    \centering
    \includegraphics[width=0.8\textwidth]{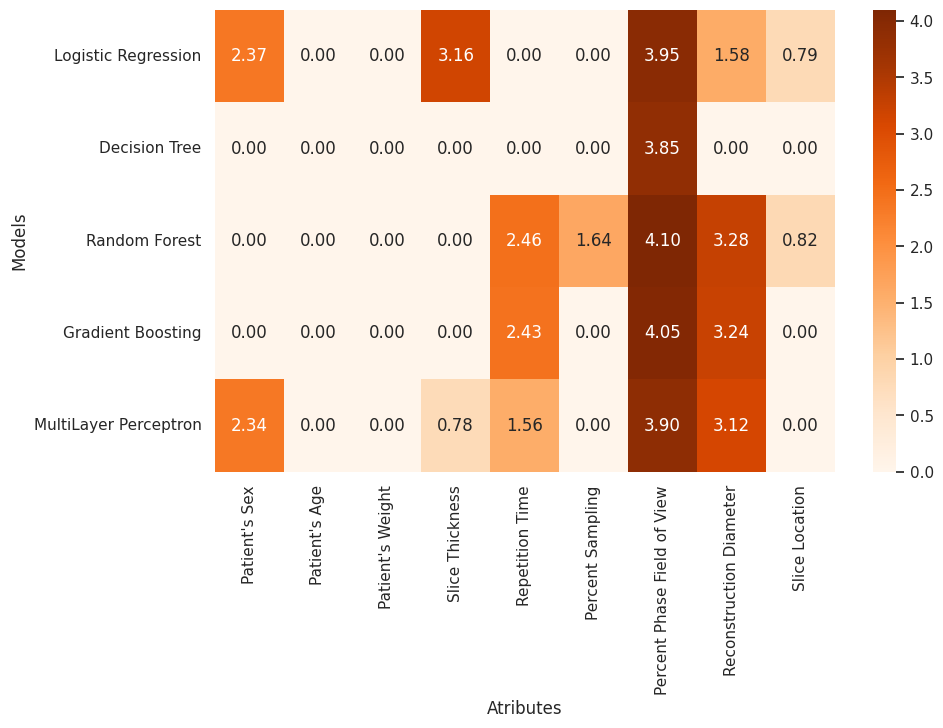}
    \caption{Top five most important features for each model, weighted by their respective F1-Score results for the LS-ST1 dataset.}
    \label{fig:LS15models}
\end{figure}

Figure~\ref{fig:LS15models} provides a comprehensive overview of the most significant parameters impacting sagittal T1-weighted lumbar spine image quality across all models. This overview indicates that pFOV, FOV, and RT are the most critical features influencing image quality in this protocol.

\subsection{C-ST2 Dataset}
\label{sub:cst2}

The performance of the models based on NCV is presented in the upper section of Table~\ref{tab:CS2performance} (boxplots are provided in the Supplementary Material, Figure~3). Based on F1-score values, RF emerged as the best-performing model, achieving an F1-score of 0.83 ± 0.04, with a precision of 0.86 ± 0.07 and an accuracy of 0.83 ± 0.04. GB followed closely, with an F1-score of 0.83 ± 0.05, standing out for its recall among all models (0.82 ± 0.08).

\begin{table}[h]
\resizebox{\textwidth}{!}{%
\begin{tabular}{llcccccc}
\textbf{Model} & \textbf{Stage} &\textbf{Accuracy} & \textbf{Precision} & \textbf{Recall} & \textbf{AUC-PR} & \textbf{AUC-ROC} & \textbf{F1} \\ \cline{1-8} 
\textbf{LR} & \multirow{5}{*}{NCV}       & 0.79 ± 0.07 & 0.80 ± 0.10 & 0.80 ± 0.09 & 0.90 ± 0.04 & 0.87 ± 0.06 & 0.79 ± 0.06 \\
\textbf{DT}  &  & 0.79 ± 0.04 & 0.80 ± 0.07 & 0.79 ± 0.08 & 0.81 ± 0.05 & 0.83 ± 0.03 & 0.79 ± 0.04 \\
\textbf{RF} &    & 0.83 ± 0.04 & 0.86 ± 0.07 & 0.81 ± 0.08 & 0.91 ± 0.04 & 0.90 ± 0.03 & 0.83 ± 0.04 \\
\textbf{GB} &        & 0.82 ± 0.05 & 0.83 ± 0.07 & 0.82 ± 0.08 & 0.91 ± 0.05 & 0.91 ± 0.05 & 0.82 ± 0.05 \\
\textbf{MLP} &      & 0.79 ± 0.07 & 0.82 ± 0.11 & 0.78 ± 0.11 & 0.90 ± 0.04 & 0.87 ± 0.06 & 0.79 ± 0.07 \\ \hline

\textbf{LR} &  \multirow{5}{*}{Test}   & 0.79 & 0.82 & 0.75 & 0.86 & 0.85 & 0.78 \\
\textbf{DT} &        & 0.83 & 0.88 & 0.78 & 0.84 & 0.85 & 0.82 \\
\textbf{RF} &        & 0.81 & 0.89 & 0.69 & 0.90 & 0.92 & 0.78 \\
\textbf{GB} &      & 0.83 & 0.90 & 0.75 & 0.89 & 0.90 & 0.82 \\
\textbf{MLP} &      & 0.88 & 0.91 & 0.83 & 0.93 & 0.93 & 0.87 \\ \hline
\end{tabular}
}
\caption{Performance metrics of the models trained on the CS-ST2 dataset, evaluated using nested cross-validation (NCV) and test data.\label{tab:CS2performance}}

\end{table}

Simpler models, such as LR and DT, as well as the more complex MLP, exhibited lower overall performance. However, when analyzing the results in the lower section of Table~\ref{tab:CS2performance}, which presents the performance of models optimized through hyperparameter tuning (detailed in Table~3 of the supplementary material), it becomes evident that RF and GB did not maintain their generalization capability on the holdout test data. This is reflected in a drop of 0.07–0.11 in their average recall values, suggesting that these models did not adequately learn the behavior of the positive class, keeping a good sensitivity in image quality classification.

While LR and DT showed improved performance on test data, as reflected by increased AUC-PR values, their recall also decreased, indicating a similar learning bias to that observed in RF and GB. On the other hand, MLP demonstrated the highest capacity to generalize from NCV, as it exhibited overall improvement in its evaluation metrics when applied to test data, surpassing the best-performing model from NCV.

\begin{figure}[!h]
     \centering
     \begin{subfigure}[b]{0.47\textwidth}
         \centering
         \includegraphics[width=\textwidth]{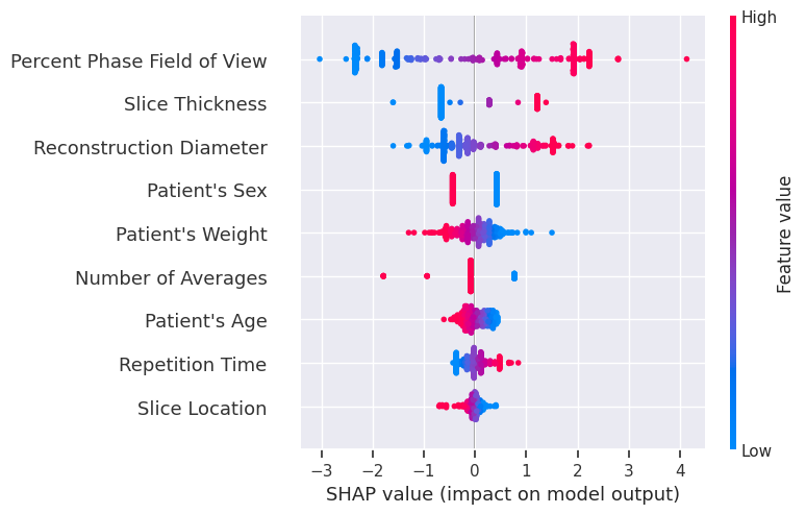}
         \caption{LR}
         \label{fig:CS2shap:LR}
     \end{subfigure}
     \hfill
     \begin{subfigure}[b]{0.47\textwidth}
         \centering
         \includegraphics[width=\textwidth]{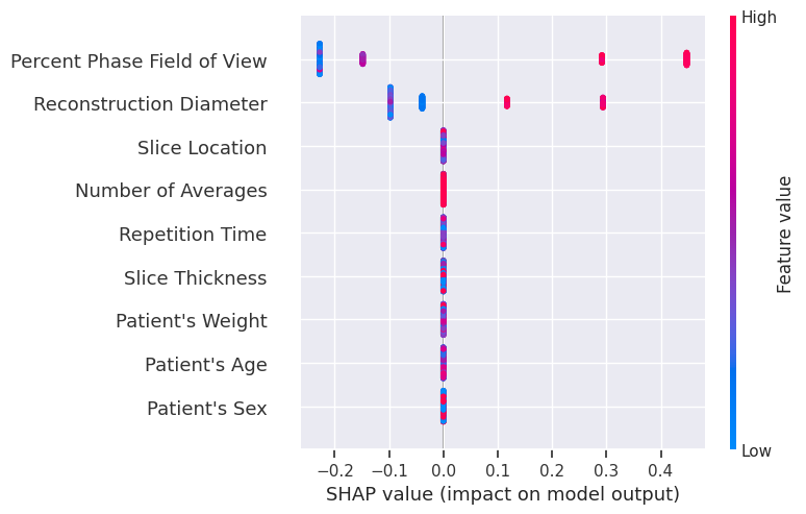}
         \caption{DT}
         \label{fig:CS2shap:DT}        
     \end{subfigure}
     \\
     \begin{subfigure}[b]{0.47\textwidth}
         \centering
         \includegraphics[width=\textwidth]{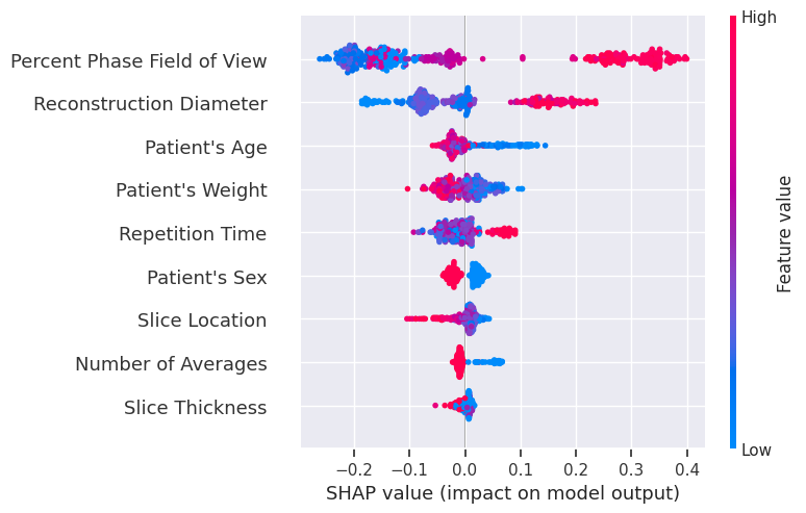}
         \caption{RF}
         \label{fig:CS2shap:RF}
     \end{subfigure}
     \hfill
     \begin{subfigure}[b]{0.47\textwidth}
         \centering
         \includegraphics[width=\textwidth]{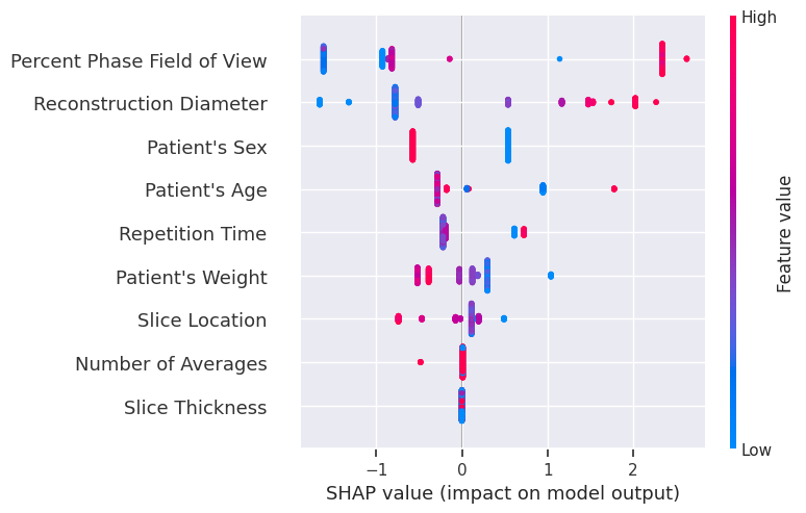}
         \caption{GB}
         \label{fig:CS2shap:GB}        
     \end{subfigure}
      \\
     \begin{subfigure}[b]{0.47\textwidth}
         \centering
         \includegraphics[width=\textwidth]{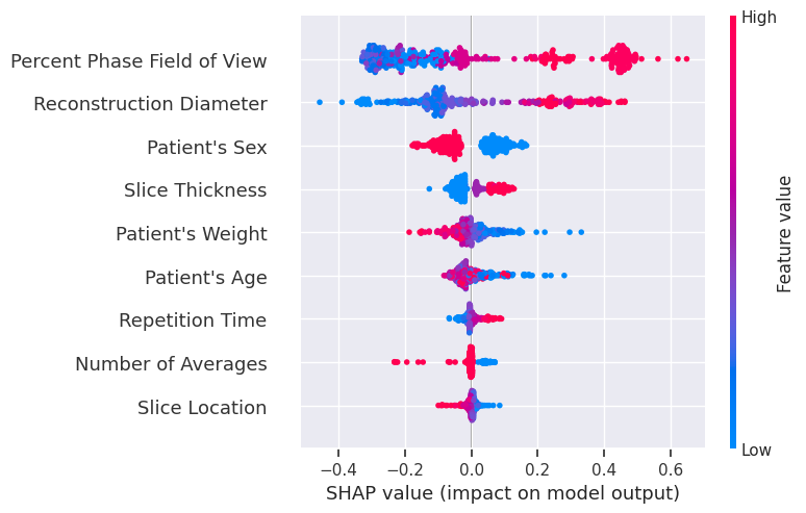}
         \caption{MLP}
         \label{fig:CS2shap:MLP}
     \end{subfigure}
     \hfill
\caption{SHAP values results for the features in the C-ST2 dataset.}
\label{fig:CS2shap}
\end{figure}

The SHAP analysis results, shown in Figure~\ref{fig:CS2shap}, indicate that for the model with the highest generalization ability in the test data, MLP, an increase in examination parameters such as pFOV, FOV, and slice thickness contributed positively to image quality classification. This way, for this protocol, in agreement with other models, slice thickness exhibited an inverse trend as observed for T1-weighted images. 

Meanwhile, pFOV and FOV consistently ranked among the top three most relevant parameters, showing a similar impact on image quality across all models, as well as in T1-weighted images. The RT trend persisted, though it exhibited lower relevance in T2-weighted protocols. Additionally, a slight trend toward lower NEX values contributing to positive image quality classification was also observed in the RF and MLP results.

\begin{figure}[!ht]
    \centering
    \includegraphics[width=0.8\textwidth]{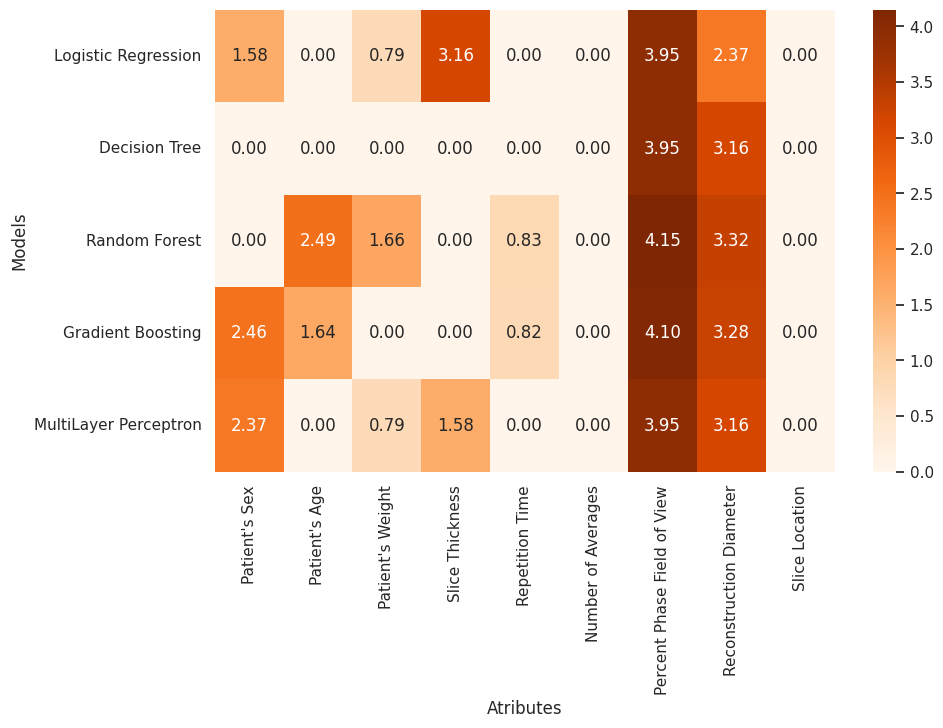}
    \caption{Top five most important features for each model, weighted by their respective F1-Score results for the C-ST2 dataset.}
    \label{fig:CS25models}
\end{figure}

Finally, Figure~\ref{fig:CS25models} highlights the consistent importance of pFOV and FOV in determining image quality. However, instead of RT, slice thickness emerged as the most influential examination parameter in classification outcomes.

\subsection{LS-ST2 Dataset}
\label{sub:lsst2}

The performance of the models on LS-ST2, the smallest dataset, as evaluated using NCV, is summarized in Table~\ref{tab:LS2performance}. We may observe lower overall performance and greater variability compared to the previously analyzed datasets. This is emphasized by the analysis of the boxplots for the computed scores, provided in the Supplementary Material, Figure~4.

\begin{table}[h]
\resizebox{\textwidth}{!}{%
\begin{tabular}{llcccccc}
\textbf{Model} & \textbf{Stage} &\textbf{Accuracy} & \textbf{Precision} & \textbf{Recall} & \textbf{AUC-PR} & \textbf{AUC-ROC} & \textbf{F1} \\ \cline{1-8} 

\textbf{LR} & \multirow{5}{*}{NCV}  & 0.73 ± 0.06 & 0.73 ± 0.08 & 0.73 ± 0.12 & 0.82 ± 0.10 & 0.80 ± 0.10 & 0.73 ± 0.07 \\
\textbf{DT} &   & 0.71 ± 0.10 & 0.67 ± 0.08 & 0.83 ± 0.14 & 0.65 ± 0.09 & 0.71 ± 0.10 & 0.74 ± 0.10 \\
\textbf{RF} &    & 0.70 ± 0.07 & 0.67 ± 0.07 & 0.80 ± 0.11 & 0.77 ± 0.09 & 0.75 ± 0.08 & 0.72 ± 0.07 \\
\textbf{GB} &         & 0.68 ± 0.08 & 0.65 ± 0.08 & 0.78 ± 0.11 & 0.68 ± 0.09 & 0.72 ± 0.08 & 0.71 ± 0.07 \\
\textbf{MLP} && 0.71 ± 0.11 & 0.71 ± 0.12 & 0.73 ± 0.14 & 0.79 ± 0.13 & 0.76 ± 0.14 & 0.72 ± 0.11 \\ \hline
\textbf{LR} &  \multirow{5}{*}{Test}   & 0.56 & 0.57 & 0.50 & 0.63 & 0.60 & 0.53 \\
\textbf{DT} &        & 0.56 & 0.55 & 0.71 & 0.53 & 0.56 & 0.62 \\
\textbf{RF} &        & 0.62 & 0.65 & 0.54 & 0.66 & 0.71 & 0.59 \\
\textbf{GB} &      & 0.56 & 0.55 & 0.71 & 0.64 & 0.68 & 0.62 \\
\textbf{MLP} && 0.58 & 0.59 & 0.54 & 0.66 & 0.63 & 0.57 \\ \hline

\end{tabular}
}
\caption{Performance metrics of the models trained on the LS-ST2 dataset, evaluated using nested cross-validation (NCV) and test data.\label{tab:LS2performance}}

\end{table}

Despite this variability, when considering the F1-score values in the upper section of Table~\ref{tab:LS2performance}, the DT model achieved the highest F1-score of 0.74 ± 0.10, primarily due to its recall of 0.83 ± 0.14. However, DT also presented the lowest precision (0.67 ± 0.08) and AUC-PR (0.65 ± 0.65) among all models. In contrast, LR obtained the second-highest F1-score (0.73 ± 0.07) while demonstrating the most consistent performance across all quality metrics, with recall values of 0.73 ± 0.12 and AUC-PR of 0.82 ± 0.10, for example.

The lower section of Table~\ref{tab:LS2performance} presents the performance metrics of models with tuned hyperparameters (detailed) in Table~4 of the supplementary material) on the holdout test data. A general trend of overfitting is evident across all models, with test performance approaching that of a random classifier. While identifying a single superior model is challenging, certain performance patterns stand out. For instance, RF achieved the highest precision (0.65) but suffered from low recall (0.54) on the test data. Conversely, GB and DT demonstrated the highest recall (0.71), albeit with the lowest precision (0.55).

\begin{figure}[!h]
     \centering
     \begin{subfigure}[b]{0.47\textwidth}
         \centering
         \includegraphics[width=\textwidth]{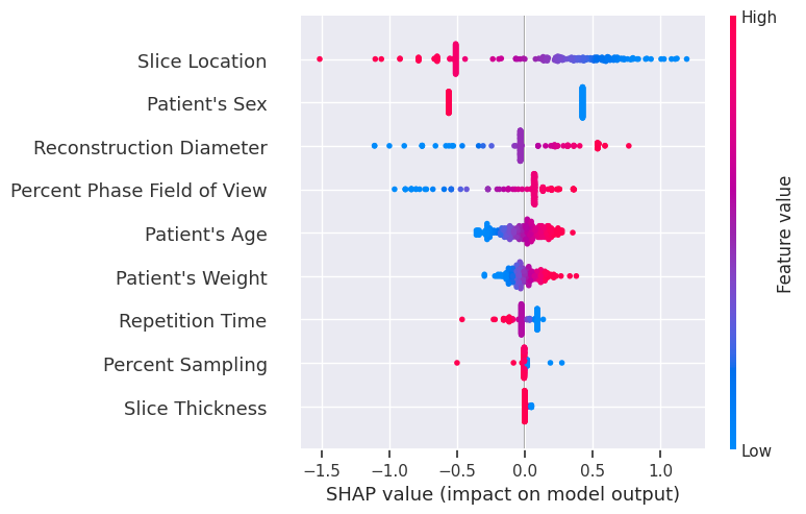}
         \caption{LR}
         \label{fig:LS2shap:LR}
     \end{subfigure}
     \hfill
     \begin{subfigure}[b]{0.47\textwidth}
         \centering
         \includegraphics[width=\textwidth]{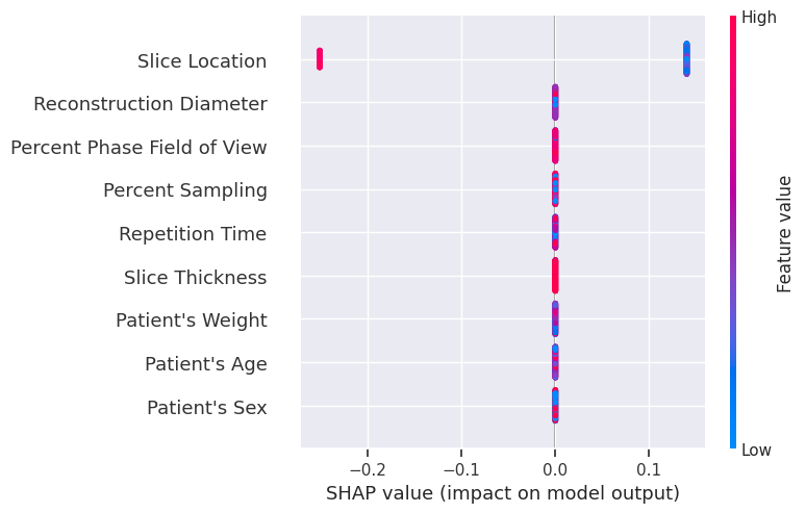}
         \caption{DT}
         \label{fig:LS2shap:DT}        
     \end{subfigure}
     \\
     \begin{subfigure}[b]{0.47\textwidth}
         \centering
         \includegraphics[width=\textwidth]{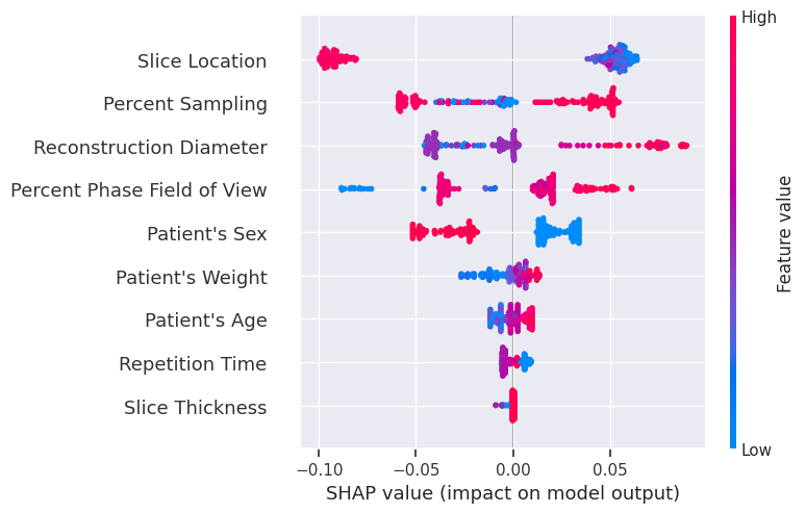}
         \caption{RF}
         \label{fig:LS2shap:RF}
     \end{subfigure}
     \hfill
     \begin{subfigure}[b]{0.47\textwidth}
         \centering
         \includegraphics[width=\textwidth]{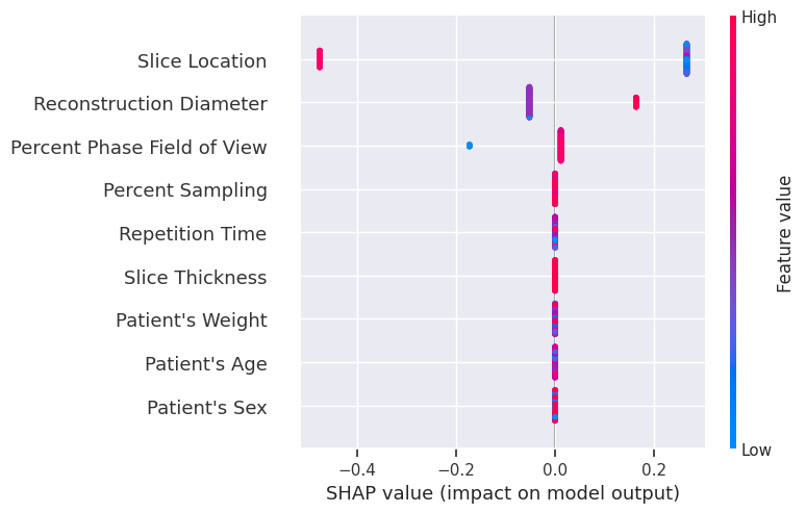}
         \caption{GB}
         \label{fig:LS2shap:GB}        
     \end{subfigure}
      \\
     \begin{subfigure}[b]{0.47\textwidth}
         \centering
         \includegraphics[width=\textwidth]{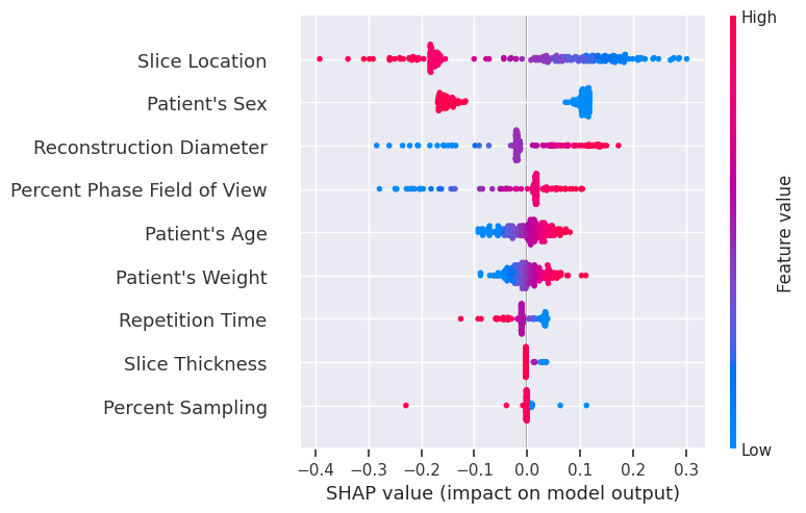}
         \caption{MLP}
         \label{fig:LS2shap:MLP}
     \end{subfigure}
     \hfill
    
\caption{SHAP values results for the features in the LS-ST2 dataset.}
\label{fig:LS2shap}
\end{figure}

The SHAP value analysis presented in Figure~\ref{fig:LS2shap} highlights the key features influencing image quality classification. In the LR model, FOV and pFOV were the most relevant examination parameters, consistently showing a positive correlation with image quality classification—higher values increased the likelihood of a positive classification. Similarly, for RF and GB, these attributes ranked among the top four most influential features. However, in RF, percent sampling emerged as the second most important feature, though the impact of the percent sampling value variation on the image quality classification was not clearly defined.

Additionally, all models identified slice location—a continuous and variable parameter related to patient positioning—as the most relevant feature for image quality classification. The results indicate that a negative displacement relative to the image center was associated with a higher probability of positive image quality classification. Subtle trends related to RT were also observed in LR, RF, and MLP models. In this dataset, a reduction in RT was associated with an increased likelihood of positive image quality classification—contrary to trends observed in all previously analyzed datasets.

\begin{figure}
    \centering
    \includegraphics[width=0.8\textwidth]{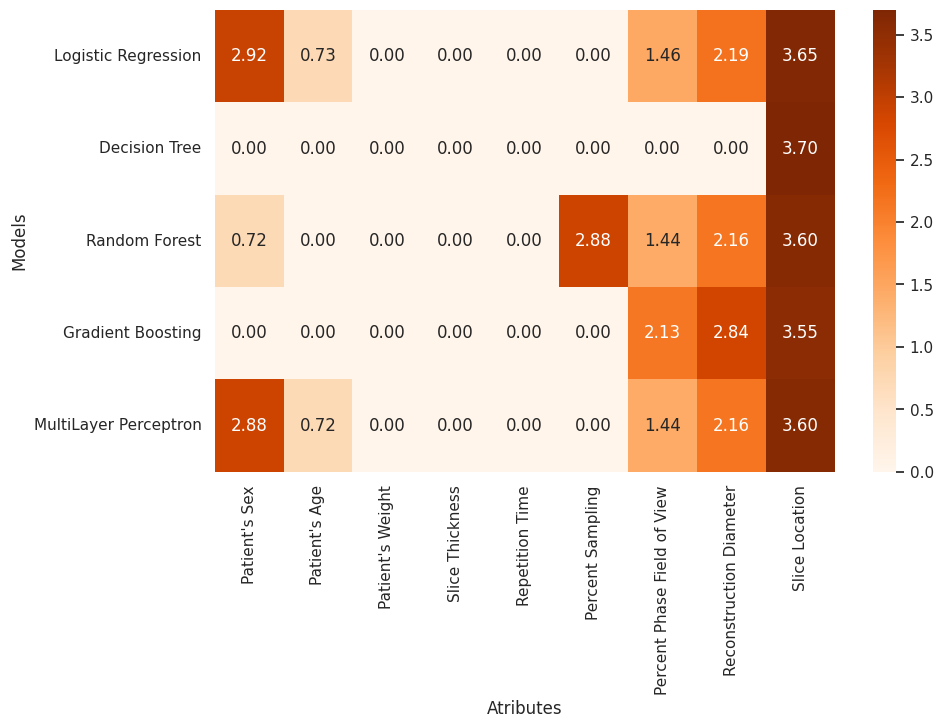}
    \caption{Top five most important features for each model, weighted by their respective F1-Score results for the LS-ST2 dataset.}
    \label{fig:LS25models}
\end{figure}

Finally, as illustrated in Figure~\ref{fig:LS25models}, the influence of slice location, pFOV, and FOV remains consistent across all models. The only other relevant examination parameter observed was percent sampling.

\section{Discussion} 
\label{sec:discussion}
\subsection{Results and Background}
\label{sub:resultsBg}

This subsection interprets the results shown in Section~\ref{sec:results} based on the theory presented in Section~\ref{sec:background}.

\begin{figure}
    \centering
    \includegraphics[width=1\textwidth]{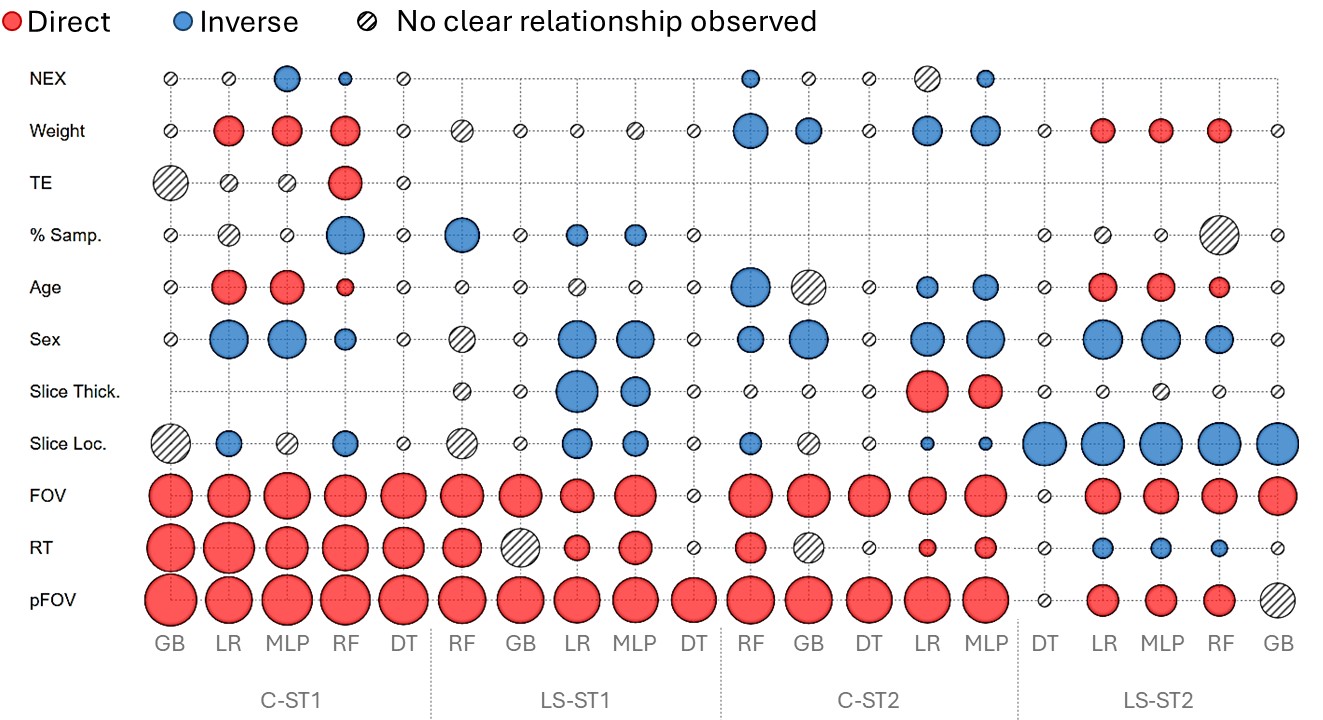}
    \caption{Results of all attributes by model and dataset, with bubble size representing the attribute's impact and color representing the trend in image quality. Red indicates a directly proportional relationship, blue indicates an inversely proportional relationship (similar to SHAP plots), and gray indicates no observable trend.}
    \label{fig:allInfo}
\end{figure}

To summarize the results for the discussion, Figure~\ref{fig:allInfo} was constructed. This figure presents the performance of all models across the evaluated datasets (horizontal axis) in relation to the selected parameters (vertical axis). At each model-dataset intersection, bubbles represent the influence of the respective parameter on the classification function, as determined by its SHAP importance ranking, weighted by the F1 training performance for that specific model-dataset pair. The size of each bubble reflects the magnitude of this impact, while the color indicates the observed relationship: red for parameters with a direct positive relationship (improved image quality with an increase), blue for an inverse relationship (improved quality with a decrease), and gray for parameters with no clear trend. Absence of a bubble at an intersection signifies that, after dimensionality reduction, the parameter was not identified as relevant for the model application.

Regarding modifiable attributes, the following were observed: NEX, TE, Sampling Percentage (\% Samp), Slice Thickness (Slice Thick.), FOV, RT, and pFOV. For the subsequent discussion, only trends agreed upon by at least two models were considered. 

Notably, the pFOV and FOV parameters exhibited consistent impacts and trends across all datasets. These parameters require manual configuration by technologists, which explains their variability and relevance. Reducing the FOV limits the amount of information used for image formation, while increasing it typically provides more information and enhances the SNR. Similarly, increasing the pFOV primarily improves SNR but also prolongs scan time.

Regarding RT, it was observed that this attribute is generally lower for T1-weighted images and higher for T2-weighted images. Moreover, as RT increases, the magnitude of the detected signal increases, resulting in a higher SNR. For T1 decay, increasing RT leads to a greater impact on the signal's intensity, which aligns with the observed impact and trend in the T1-weighted datasets. However, excessive increases in RT can compromise the T1 contrast. This suggests that contrast may need to be handled with a different methodology.

In T2-weighted images, the impact of RT generally aligns with theoretical expectations, except in the LS-ST2 dataset, where the results contradict anticipated trends. This discrepancy might stem from the limited sample size of the LS-ST2 dataset, restricting the models' capacity to learn effectively.

The decrease in signal amplitude after RT is influenced by the loss of coherence between spins and the external magnetic field. Consequently, T2-weighted images with longer TE exhibit lower overall signal intensity compared to T1-weighted images. This characteristic may explain why slice thickness shows opposing trends in T1- and T2-weighted datasets.

The impact of slice thickness on image quality is ambiguous. Images with thicker slices generally have higher SNR but poorer spatial resolution, while thinner slices result in lower SNR but better spatial resolution \cite{parametersBkg}. The trend of the slice thickness attribute can thus indicate whether image quality measurements prioritize spatial resolution or SNR.

This ambiguity is evident in Figure~\ref{fig:allInfo}, where slice thickness trends were observed in the C-ST2 and LS-ST1 datasets. The findings suggest that the signal loss associated with thinner slices in T2-weighted images may have a greater influence on image quality compared to T1-weighted images, which tend to benefit more from higher resolution despite the loss of signal. This observation aligns with theoretical expectations and is supported by average dataset values, as technologists often prefer thicker slices for T2-weighted images. These results demonstrate the capacity of the models to pragmatically learn theoretical principles.

Lastly, the trends observed in NEX values and sampling percentages suggest that smaller values increase the likelihood of positive classification for image quality. Excessive reduction in NEX and sampling percentage may result in poor-quality images \cite{parametersBkg}, but a subtle reduction can help decrease motion artifacts (blurring), which is particularly relevant in spine exams \cite{nexSpine}.

Thus, by applying the proposed methodology to DICOM metadata, it was possible to identify parameters with the most variability during an exam, such as pFOV and FOV in spine exams. These parameters require manual configuration by MRI technologists, and their variability underscores the importance of further studies aimed at standardizing these attributes for optimized protocols. Furthermore, it was observed that the methodology is capable of learning from technologists about the foundational rules of MRI protocols. This capability enables the identification of practical service limitations and provides tailored suggestions for equipment-specific protocol optimization. 
It is important to note that final quality measures should consider the reduction of acquisition time in order to provide a model with practical utility.

\subsection{Study Limitations}
\label{sub:limitations}

While the findings are promising, the study acknowledges several limitations. First, the performance evaluation methodology, which relied on test data from the holdout method, poses a potential constraint on the generalizability and robustness of the models. Although nested cross-validation was employed for hyperparameter tuning and model selection, the limited sample size makes the holdout approach highly susceptible to biases or specific characteristics of the dataset, potentially impacting the results and reliability of performance evaluation.

Moreover, the selected quality measures were not directly aligned with expert assessments. While the study prioritized improvements in SNR and spatial resolution, it did not adequately address changes in image contrast, which is a critical aspect of image quality.

Another limitation lies in the models' performance when dealing with smaller sample sizes, particularly in the LS-ST2 dataset, which contained fewer than 292 images. This underscores the need for alternative strategies, such as ensemble methods, to enhance model reliability in scenarios with limited data availability.

\subsection{Future Works}
\label{sub:future}
This study was not intended to construct a fully deployable, practical model, as its primary focus was on implementing and experimentally evaluating the proposed ML-based methodology to optimize MRI protocols, particularly given the lack of previous studies supporting the feasibility of such an approach. Therefore, future work should include the incorporation of acquisition time as a component of the image quality measure. This addition will be crucial for developing a practical, deployable model and ensuring its subsequent validation.

Moreover, while image quality measures are valuable, they should not replace expert opinions, though they can provide useful indications. Consequently, methodologies such as the one presented in \citet{optimStudy} should be explored when constructing quality-targeted attributes. It is critical that this methodology is grounded in models capable of continuous learning, considering the evolving characteristics of MRI equipment over time. While the current measure aligns with SNR, as detailed in Subsection~\ref{sub:resultsBg}, exploring additional metrics for other important characteristics -- such as spatial resolution, contrast, or even combinations of multiple metrics -- would further enhance the robustness and generalizability of the methodology.

Additionally, based on the results from the LS-ST2 dataset, it is clear that for smaller sample sizes (fewer than 292 DICOM images), alternative methodologies, such as ensemble learning, should be considered. The models derived from this dataset exhibit diverse performance strengths that could be combined to improve overall performance. For example, combining the RF model (which showed higher precision) with the GB model (which demonstrated better recall) could lead to more balanced performance and effective predictions. 

\section{Conclusion}
\label{sec:conclusion}

This study successfully identified the consistency of trends for key parameters -- including pFOV, FOV, TR (with the exception of the LS-ST2 dataset), slice thickness, sampling percentage, and NEX -- in relation to the practical variations and their impact on quality measures derived from spectral flatness and entropy power. These findings validate the feasibility of the proposed methodology in learning from real-world data and suggesting informed modifications to protocol quality parameters, particularly in optimizing image quality, with a focus on improving SNR.

The ability to develop a model capable of dynamically identifying parameter trends for image optimization, grounded in practical considerations, ensures the robustness and applicability of MRI quality improvement programs. Furthermore, the methodology's potential for ongoing learning, in parallel with advancements in technology, not only contributes to improving MRI protocol optimization but also offers valuable insights that can enhance theoretical understanding in the field.

As highlighted by Einstein, ``\textit{In theory, theory and practice are equal. In practice, they are not.}" This underscores the importance of bridging theoretical concepts with real-world applications, particularly in the context of medical imaging. Continuous efforts to align technological innovations with the practical realities of MRI processes are crucial for advancing healthcare outcomes and ensuring the effectiveness of medical imaging protocols.

\section*{Ethics declaration}
This study was conducted using anonymized MRI data collected from the Hospital de Clínicas de Porto Alegre (HCPA). All data were de-identified to ensure patient confidentiality and were handled in accordance with ethical standards.  The study was submitted to the Research Ethics Committee of Hospital de Clínicas de Porto Alegre and it was approved under number 2023-0290 (CAAE 74933423.2.0000.5327), ensuring compliance with local regulations and ethical guidelines for the use of medical data in research. No personally identifiable information was used or disclosed during the course of this study, and the data were accessed with the appropriate institutional permissions.

\section*{Consent for publication}
This manuscript does not require consent for publication as it involves secondary data analysis, and no personally identifiable information is included. 

\section*{CRediT authorship contribution statement}
\textbf{Alice Vian:} Conceptualization, Data Curation, Methodology, Formal analysis, Investigation,  Writing - Original Draft, Visualization. \textbf{Mariana Mendoza:} Conceptualization, Methodology, Validation, Writing - Review \& Editing, Supervision, Project administration. \textbf{Diego Eifer:} Resources, Writing - Review \& Editing, Supervision. \textbf{Mauricio Anes:} Resources, Writing - Review \& Editing. \textbf{Guilherme Ribeiro Garcia:} Resources, Writing - Review \& Editing.

\section*{Conflict of interest}
The authors declare that they have no conflict of interest.

\section*{Funding}
This study was financed in part by the Coordenação de Aperfeiçoamento de Pessoal de Nível Superior - Brasil (CAPES) - Finance Code 001, and by grants from the Fundação de Amparo à Pesquisa do Estado do Rio Grande do Sul - FAPERGS [22/2551-0000390-7 (Project CIARS)] and Conselho Nacional de Desenvolvimento Científico e Tecnológico (CNPq) [308075/2021-8].

\section*{Declaration of generative AI and AI-assisted technologies in the writing process}
During the preparation of this work the authors used OpenAI's language model to assist in reviewing, editing, and enhancing the readability and language of this manuscript. After using this tool, the authors reviewed and edited the content as needed and take full responsibility for the content of the published article.

\section*{Acknowledgments}

We are grateful to the Radiology Service and the Medical Physics Service of Hospital de Clínicas de Porto Alegre for granting access to their data, and for their expertise and time. 

\appendix
\section{Supplementary materials}
\label{app1}
Supplementary material associated with this article can be found in the online version.

\bibliographystyle{unsrtnat}



\end{document}